# Trust at Your Own Peril: A Mixed Methods Exploration of the Ability of Large Language Models to Generate Expert-Like Systems Engineering Artifacts and a Characterization of Failure Modes

**Taylan G. Topcu[1], PhD**
Assistant Professor
Grado Department of Industrial and Systems Engineering, Virginia Tech
1145 Perry St., Blacksburg VA 24061
ttopcu@vt.edu

**Mohammed Husain**
Solutions Architect
Scale AI
155 5th St, San Francisco, CA 94103
mhusain01@vt.edu

**Max Ofsa**
**PhD Student**
Grado Department of Industrial and Systems Engineering, Virginia Tech
1145 Perry St., Blacksburg VA 24061
jaster15@vt.edu

**Paul Wach, PhD**
Research Assistant Professor
Virginia Tech National Security Institute
1311 Research Center Dr, Blacksburg, VA 24060
paulw86@vt.edu

---

[1] Corresponding Author: Dr. Taylan G. Topcu, ttopcu@vt.edu

**ABSTRACT**

Multi-purpose Large Language Models (LLMs), a subset of generative Artificial Intelligence (AI), have recently made significant progress. While expectations for LLMs to assist systems engineering (SE) tasks are paramount; the interdisciplinary and complex nature of systems, along with the need to synthesize deep-domain knowledge and operational context, raise questions regarding the efficacy of LLMs to generate SE artifacts, particularly given that they are trained using data that is broadly available on the internet. To that end, we present results from an empirical exploration, where a human expert-generated SE artifact was taken as a benchmark, parsed, and fed into various LLMs through prompt engineering to generate segments of typical SE artifacts. This procedure was applied without any fine-tuning or calibration to document baseline LLM performance. We then adopted a two-fold mixed-methods approach to compare AI generated artifacts against the benchmark. First, we quantitatively compare the artifacts using natural language processing algorithms and find that when prompted carefully, the state-of-the-art algorithms cannot differentiate AI-generated artifacts from the human-expert benchmark. Second, we conduct a qualitative deep dive to investigate *how* they differ in terms of quality. We document that while the two-material appear very similar, AI generated artifacts exhibit serious *failure modes* that could be difficult to detect. We characterize these as: premature requirements definition, unsubstantiated numerical estimates, and propensity to overspecify. We contend that this study tells a cautionary tale about why the SE community must be more cautious adopting AI suggested feedback, at least when generated by multi-purpose LLMs.

**Keywords:** systems engineering; generative artificial intelligence (AI); large language models (LLMs); human-AI collaboration; problem formulation.

## 1. Introduction

The old saga of complex system development continues as the vast majority of government and industry programs continue to result in cost and schedule overruns [1–7]. While numerous researchers and government reports attribute this trend to the ever increasing complexity of the systems being developed [8,9], future systems will only need to be more interoperable[10] and intelligent[11]; which could exacerbate the typical shortcomings of development programs. Additionally, the interdisciplinary nature of these development tasks exceed the information processing capability of any individual designer or disciplinary engineering team[12,13]. Thus, effective and efficient management of the collaborative development effort will continue to lie at the crux of the problem.

Given this context, the discipline of Systems Engineering (SE) relies on diverse teams of subject matter experts [14,15] that iteratively exchange design information and domain knowledge over extended periods of time[16]. The challenge of the systems engineer is to govern the activities of these large interdisciplinary teams by employing an array of sociotechnical skills[17,18] through effective problem formulation[19], solving[20], verification[21,22], and validation[23]. This necessitates tacit knowledge from multiple domains to be synthesized[24,25], along with high familiarity with the specific context of the system of interest[26].

Reliance on large number of experts from multiple disciplines causes an inherent disconnect among design artifacts created along the process. To elaborate, individual experts capture their design information in mutually exclusive models (or languages) such as CAD drawings, multi-physics models, various simulations, or SysML models. These collectively convey the necessary design information; however, individually, only some aspect of the whole[13]. This creates a natural gap between collaborators. Furthermore, as new design information becomes available in one of these interrelated development artifacts, changes need to be propagated to others[27]. This in return leads to additional communication workload[28], rework[29], and ultimately, schedule and cost overruns[30]. While digital transformation[31] (DT) aims to synergistically integrate these "siloed" models in a digital ecosystem to bridge the gap between collaborators[30] and ultimately expedite the SE process[32], roughly 90% of DT efforts are documented to result in failures[33,34]. There is a pressing need for rapid, data-friendly, and efficient methods for both improving SE outcomes and facilitating DT[11,35].

To that end, generative Artificial Intelligence (AI), such as large language models (LLMs), bear great promise as they have the potential to dramatically alter the way design information is represented, managed, and created[36]. SE literature has been exploring the use of broader AI applications for assisting designers; however, the focus is on formulation of knowledge databases [37,38], virtual assistants [39–41], and design evaluators [42,43]. These are documented to be helpful for improving designer performance and learning [44–46]; however, these are mostly custom-built applications instead of multi-purpose LLMs that are trained on data that is broadly available on the internet. Research on the efficacy of LLM methods to assist SE tasks, not only in terms of providing a proof of concept but achieving a desired quality threshold that would be expected of a human SE expert, is nascent[47].

This study addresses two time-sensitive objectives in pursuit of the aforementioned overarching goal. First, we extend our earlier work and provide the existence proof for the ability of multi-purpose LLMs that are trained on broadly available internet data to generate human expert-like SE artifacts[48]. We achieve this by picking a text-rich and relatively complete human-expert generated SE case study as a benchmark, and then select a set of LLMs to experiment with different prompting strategies to generate SE artifacts for the system of interest in the benchmark. We then adopt a mixed-methods approach, and compare the AI generated artifacts against the benchmark through both quantitative and qualitative methods. Here, the quantitative perspective pertains to comparing AI generated vs. human-expert benchmark by using a semantic natural language processing algorithm that measures similarity[49]. From this lens, we find that LLMs can generate SE artifacts that closely resemble those generated by human-experts; however, the prompting approach has a significant influence over the quality of outputs.

Second, we complement this analysis with a qualitative analysis of the AI generated results, to (i) explain *how* AI generated artifacts differ from the human-expert benchmark in terms of quality and (ii) document in *which aspects they are comparable*. We find that even the most similar AI generated artifacts exhibit serious deficiencies, which we characterize around three *failure modes*. Namely, these are: *premature requirements definition*, *unsubstantiated numerical estimates*, and *propensity to overspecify*. While we observe these failure modes in the context of the system of interest used in the benchmark, the rich comparison against the human-expert benchmark and the characteristic their patterns of emergence suggest that these failure modes may be *generalizable* to other LLM use cases regarding SE tasks.

Collectively, findings of this paper should be taken as a conservative estimate of AI capabilities for SE, while also telling a cautionary tale of how misleading these artifacts could be particularly when relying on models that are trained on the broader internet instead of a proprietary database or expert knowledge. AI generated artifacts that appear well constructed, or even unintelligible from the lens of a novice or from a language processing perspective, may be seriously misleading and could lead to catastrophic design decisions with cascading effects. This calls for additional research into verification and validation of AI generated artifacts for SE, along with an increased attention on development of customized AI that is contextualized for specific SE tasks[50–52]. That being said, this is a field that is advancing rapidly, as such, LLMs examined in this study have already been replaced with newer, more capable ones. To that end, this paper should be treated as a benchmark of the state of the art at the time, and it remains to be seen to which extent these findings will remain relevant.

The rest of this article is organized as the following. Section 2 provides an overview of the relevant literature. Section 3 outlines the methodology. Section 4 presents the findings from quantitative and qualitative perspectives. Section 5 provides a discussion of the findings along with implications for the field, and concludes.

## 2. RELATED LITERATURE

### 2.1. Large Language Models

#### 2.1.1. A Short History

In this section, we provide a brief history and an overview of LLMs by describing how they function. LLMs originate from the field of Natural Language Processing (NLP), an interdisciplinary field that lies on the intersection of computer science and linguistics that aims to render natural human language palatable to computers. NLP achieves this through executing a combination of tasks such as information retrieval, text classification, and language generation[53]. Earliest versions of NLP were predominantly rule-based[54]. These lacked contextual awareness but were highly-domain specific[55]. Rule-based models were followed by a transition towards learning-based methods that rely on statistical inference[56] to predict the sequence of words, such as n-gram models[57], which were later followed by hidden Markov models[58,59]. These models were powerful in terms of speech recognition and information retrieval, yet struggled with the ambiguity, particularly for handling long range dependencies.

The new era of NLPs was enabled through the use of deep learning methods. Initially, artificial neural networks provided significant benefits in terms of flexibility. However, these struggled in terms of handling sequential data[60] and were later replaced by recurrent neural networks that are successful in capturing sequences over time through backpropagation methods [61]. Later, a special instance of recurrent neural networks, long-term short-term memory (LSTM) models were developed. These achieved a significant breakthrough in NLP performance by selectively retaining information for longer periods of time[62,63]. Nevertheless, it wasn't until the introduction of transformer models that NLPs exhibited a significant performance leap [64]. These fundamentally differ from recurrent network-based approaches as they rely on self-attention mechanisms, that allow them to process sequences of information in parallel, as opposed to the sequential approach of neural networks.

Multi-purpose LLMs have dramatically changed the landscape of NLP capabilities in the last two years[65]. While the development of LLMs have been a work in progress for years by companies such as OpenAI, Anthropic, and Google, it wasn't until the release of GPT-3 (Generative Pre-trained Transformer) by OpenAI in 2022 and the subsequent release of "ChatGPT" or GPT-3.5 in November 2023 that LLMs became commonplace names and tooling. Briefly, LLMs follow a three-step process:

- **Step 1: Pre-training:** the model is exposed to enormous text-based information and uses self-supervised learning to understand the role of words in context. Specifically, the LLM takes a snippet of text and masks a random word or token (i.e., a subset of a word), then attempts to predict that word given the context of the preceding words. Examples of publicly available pre-training datasets include CommonCrawl (12 years of web crawling), WebText (8M documents from Reddit), Wikipedia, and arXiv.

- **Step 2: Supervised fine-tuning (SFT):** the model is fine-tuned by providing it with a series of prompts and human-generated responses to complete specific tasks, such as summarization, entity extraction, sentiment analysis, translation, and text generation - i.e., prompt-response pairs can be thought of as feature-label pairs, using a classical machine learning analogy. These "prompt-response" pairs are typically generated from data labelling companies (e.g., Scale AI, Surge AI) or can also be boot-strapped using LLM responses to a series of prompts.

- **Step 3: Reinforcement learning (RL):** model responses are ranked by a series of experts, this is commonly referred to as "reinforcement learning with human feedback"[66]. These rankings are then used to train a reward model to learn implicit human preferences and encourage the model to generate responses aligned with these preferences. These preferences can either be subjective (e.g., moral, political, ethical) or objective (e.g., accuracy, length of response, quality and diversity of response).

*2.1.2. A Simplified Demonstration*

Understanding how LLMs are formulated and the training datasets used in each step is critical for the purpose of generating SE artifacts, particularly for Government entities such as the Department of Defense (DoD). Due to the massive and broad scope of the pre-training datasets, LLMs exhibit strong performance at reasoning over a spectrum of different domains thus can yield "off-the-shelf" value across many use cases prior to any domain-specific supervised fine-tuning model. To illustrate, Figure 1 presents a LLM-generated response to a basic SE question. While one could reasonably consider the response provided in Figure 1 as appropriate and perhaps high-quality, there is clearly room for improvement and an opportunity to "tailor" it for a given end-user.

```
Input: What is an OV-1 diagram?

Output: An OV-1 diagram, or Operational View-1 diagram, is a high-level graphical representation
used in systems engineering to illustrate the operational objectives, capabilities, and operational
environment of a system. It provides a broad overview of the mission, relationships between systems,
and the interactions with the environment.
```
**Fig. 1.** Example of LLM inference on a SE use case

Thus, to tailor the LLM response, at least for ones that are trained on the public data on the internet, the missing piece is the concept of "prompt engineering"[67]. Prompt engineering is a relic of the massive corpus of pre-training data and alludes to an extremely high degree of sensitivity the model has to the provided inputs. By providing an LLM with a more specific prompt, either stylistically (e.g., specifying the length and style of the response) or contextually (e.g., specifying a particular domain or persona for whom to tailor the response), LLMs can provide higher quality responses. For example, Figure 2 revisits the same question, prompted for a DoD specific persona. Note that the response has improved in quality by using domain-specific vernacular, supplies helpful use cases for the artifact, and even adheres well to the syntactic requirement imposed in the prompt. Here, the contrast between Figures 1 and 2 illustrate how the LLM can tailor its response based on given prompts, while providing general SE related information that could be useful for SEs that may not be familiar with US DODAF framework.

```
Input: What is an OV-1 diagram? Focus your response on relevance to a Defense-specific audience
and keep your response to one concise but comprehensive paragraph.

Output: An OV-1 diagram, or Operational View-1 diagram, is a high-level graphical representation
used in the Department of Defense Architecture Framework (DoDAF) to depict the strategic mission,
vision, and capabilities of an organization or a military operation. It provides a broad overview
of the operational concept, including the relationships among key mission areas, organizational
structures, and external interfaces. The OV-1 diagram is crucial for defense
-specific audiences as it helps in understanding the operational needs, identifying potential
risks, and planning strategic actions in a military context. It serves as a communication tool for
conveying complex operational scenarios to a wide range of stakeholders, including military leaders,
defense planners, and system developers.
```
**Fig. 2.** How prompt engineering aligns the response to a user-specific domain and use case

The case illustrated in this subsection uses an open access LLM that is trained on publicly available data. Given that many organizations that have significant amount of proprietary and context specific data, created by domain-experts, it should be considered only a conservative demonstration of potential benefits. Thus, we contend that LLMs can play a significant role in cost-effective generation of SE artifacts due to their extensive pre-training, and largely as a function of clever prompt engineering, even in the absence of domain-specific fine-tuning.

### 2.2. Generative Artificial Intelligence for Systems Engineering

There is a growing body of literature in the broader engineering design and SE community on the use of generative AI for various aspects of the system lifecycle[36]. These could be categorized in requirements engineering, concept generation & tradespace exploration, and human-AI collaboration during decision-making in design. We consider it useful to provide a brief overview.

Requirements engineering has been the pioneering use case in SE for NLPs, and later LLMs, given the wide-spread availability of text based data[68,69]. Nevertheless, this thrust also accelerated significantly with the advancement of LLMs and currently pursues a plethora of objectives. These include defining specifications[70], classification and extraction of requirements types from other engineering documents[71,72], establishing traceability[73,74], identification of user needs[75], and knowledge extraction[51]. Recent research in this area benchmarked multi-modal LLMs for understanding reference engineering documentation, textual requirements, and CAD drawings[76]; but has also documented that there are significant shortcomings in terms of reliability and performance.

Concept generation and tradespace exploration is one of the rapidly advancing use-cases[77]. Sarica and Luo use large semantic networks for mining patent databases to aid the ideation process and illuminate novel design concepts[78]. Others experimented with generative networks for design space exploration, illustrating that both design novelty and quality could be increased concurrently by enabling access to a larger portion of the design space[79,80]. Some generative models developed in the mechanical design community pursue generation of concepts[79,81] and assisting ideation[82]. Within the SE literature, there are examples like Selva's work in the development of virtual assistants for assisting system architecture articulation for spacecraft missions[39,40,83,50].

Another research thrust is on decision-making in design, particularly on human-AI collaboration; documenting nuanced findings. Here, research is being conducted on both human-AI pairs and teams of humans with AI advisors[84,45]. There is evidence suggesting that AI assistance can concurrently improve design performance and designer learning, particularly if collaboration continued for extended periods of time[85]. However, there are also studies from a similar setting indicating AI assistance could be detrimental to designer learning[83]. Others investigated AI-assistance during design decision-making and found that designer self-confidence plays a critical role in their trust in AI, error attribution, and eventual adoption of AI guidelines[86]. Nevertheless, human trust in AI guidance is documented to depend on an array of factors, including but not limited to the accuracy of AI recommendations, transparency, fairness, and robustness[87]. Others document that AI recommendations may improve the performance of low performing teams or individuals; however, it may also hinder the performance of high performers[88,89]. At least for now, AI is not a panacea for design problem-solving[88].

## 2.3. The Gap

While there is significant research in the broader engineering design and SE literature on use of AI, there are only few examples on the use of multi-purpose LLMs that leverage historical SE artifacts to better inform the design of a new system. The most relevant to this paper is Daphne[39,40], an intelligent assistant to support preliminary system design and tradespace analysis of Earth-facing satellite systems. However, as opposed to this study, the novelty of Daphne is rooted in the purposefully trained rule-based knowledge repository and the integration of the natural language interface, instead of communication schema or protocol, which enables the ability to scale and track existing knowledge, while adjusting to new incoming data. Nevertheless, these are purposefully built models for a specific context or objective that are challenging to develop and sustain. Thus, these do not address queries regarding the efficacy of multi-purpose LLMs (e.g., Chat-GPT 4, Claude 3.5, Gemini 2.0) to generate expert-like SE artifacts, particularly without any specific fine-tuning, additional training, or customization; and when guided by a human engineer through iterative prompting.

Although this decision to not conduct any-fine tuning may seem like an odd choice, we contend that this is becoming an increasingly important query as both SE practitioners, students, and researchers are experimenting with multi-purpose LLMs as-is, without any customization or fine-tuning, to alleviate some of their workload; often with little consideration to its potential risks. Furthermore, more often than not, there is a tendency to accept these LLM outputs without further verification. We deem that this is a concerning trend and so far, there are no published studies in the SE community that have documented its potential consequences. Here, we delve into *to which extent* these multi-purpose LLMs can be effective in assisting SE tasks, and characterize *how* they fail. Thus, our aim is not to document how the best possible quality LLM outputs could be obtained, but to characterize what are the downside risks even with LLM generated artifacts that closely resemble those produced by human-experts.

## 3. METHODOLOGY

The objective of this paper is to empirically investigate if multi-purpose LLMs can generate SE artifacts that resemble those created by experts and to document if and how they may be misleading. We pursue this objective by adopting a human-expert generated SE artifact as a benchmark. We then provide various LLMs with chunks of curated data taken from this benchmark and evaluate their ability to generate SE artifacts. We evaluate AI generated artifacts through a mixed-methods approach. First, we adopt a quantitative approach and ask a machine, more specifically a specialized NLP algorithm, to test if AI generated artifacts are passable and are similar to benchmark artifacts generated by human-experts. We use this step to filter AI Generated responses in terms of similarity. Second, to compare the quality of these artifacts, we adopt a human-expert perspective and dig through AI generated artifacts through a qualitative approach to identify which extent they are similar, and *where* and *how* they diverge. Figure 3 presents the overview of our research methodology. We elaborate each of the steps below.

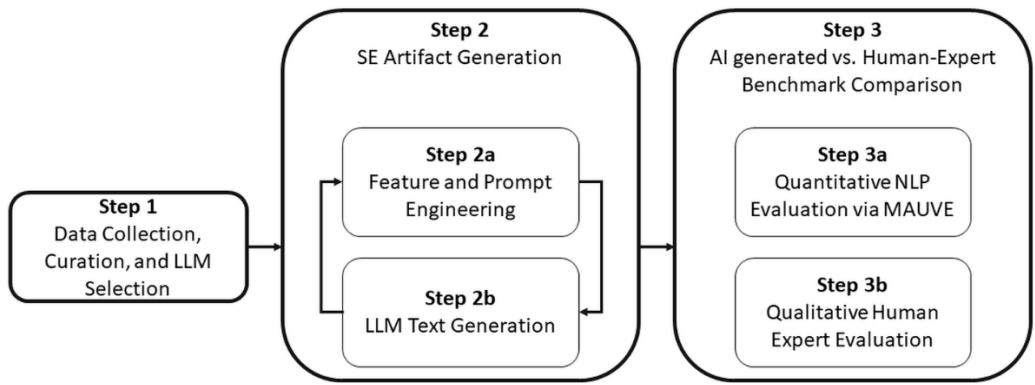

**Fig. 3.** Overview of the Research Approach

### 3.1. LLM Selection, Human-Expert Benchmark Selection and Data Processing

*3.1.1. LLM Selection*

We are interested in evaluating publicly available LLM capabilities; however, we don't have access to their design. Therefore, we choose a representative set of models to run our empirical exploration. We choose to work with exclusively closed-source models as they have the broadest performance, were trained on the most tokens, and have a large enough context window to allow a well-sized prompt. Thus, focus on the use of three closed-source models (e.g., available only via API, with no direct access to model weights): GPT-3.5 Turbo, GPT-4, and Claude. Next, we proceed to selection of the data and its pre-processing.

*3.1.2. Selection of the Sample Dataset and its Pre-processing*

The primary technical approach is to attempt to use a series of LLMs to generate SE artifacts using materials from the Bulldog Unmanned Ground Vehicle (UGV) dataset[92] corresponding to a hypothetical ACAT II program. Bulldog was conceived by the Defense Acquisition University (DAU) faculty to teach SE and is a notional UGV that is required to detect and neutralize threats (e.g., landmines). The mission set, to which Bulldog belongs, includes interfaces with command & control (C&C) through a satellite network and with an unmanned air system (UAS) that acts to detect and notify C&C of a threat presence. The UAS also acts as risk mitigation to potential loss of satellite network capability, in which case the UAS will directly notify and task the Bulldog UGV. We visualize the mission set in Figure 4 using a descriptive image, traditionally referred to as an operational view 1 (OV-1).

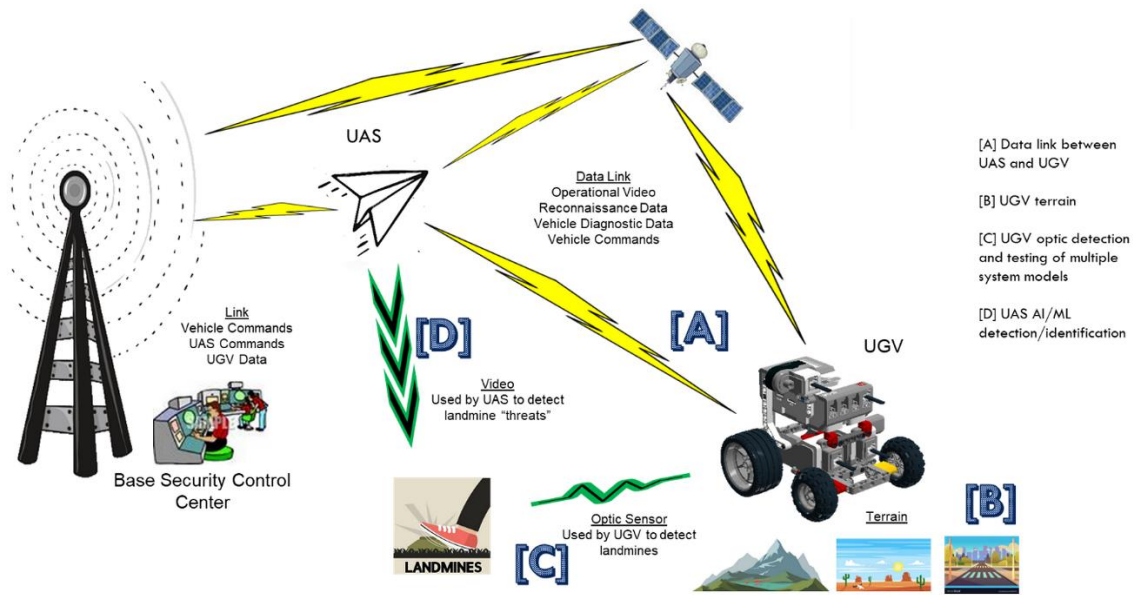

**Fig. 4** Bulldog Case Study Mission Description (OV-1)

The Bull Dog case study provides a rich and partially complete dataset. It is composed of a set of multi-page pre-Milestone B artifacts including a Capability Description Document (CDD), a System Performance Specification from a Technology Maturation and Risk Reduction (TMRR) Phase, an Initial Capabilities Document (ICD), a Systems Engineering Plan, a series of Operational Scenario Descriptions, and a Preliminary Market Research Report of Platform Requirements and Characteristics. We use this textual data and curate it for LLM processing. Therefore, our sample size for the human-expert benchmark is only equal to one; however, it has been created by a team of SEs instead of an individual. While no SE artifact is perfect, and there could possibly be other acceptable answers to a SE problem formulation question; given that this artifact was specifically developed by a team of DoD tasked SE experts, and was later independently approved for education of DoD personnel, we deem it appropriate to serve as the ground-truth for the purposes of this paper. This research choice also allows us to decouple our analysis from artifact generation; therefore, helps prevent potential biases.

*3.1.3. Data Curation*
Next is the curation of representative artifacts. For this research, we are focused on generation of full-text artifacts, as opposed to JSON objects that would be stored in a MBSE tool. The Bulldog dataset is ideal here as it contains several multi-page documents. Thus, we apply data engineering to prepare the data for ML consumption. The strategy is to chunk the 7-8 documents available in the Bulldog dataset in a series of 50 "instances," where each instance is roughly 1-2 paragraphs of text.

There are several reasons why this approach is needed. First, the context window of most LLMs is constrained to only a few thousand tokens thus require input text to be synthesized. This also helps augment the dataset to allow for greater test fidelity. Second, the MAUVE framework itself – which we use to measure similarity - requires a minimum of 50 instances to make a representative assessment. In other words, our data engineering process is shaped by our method of evaluation. In short, we distill Bulldog artifacts into a series of roughly 52 prompt-response pairs, which are then used to evaluate LLMs.

### 3.2. SE Artifact Generation Using LLMs

For each SE artifact chunk generated in this study, we start-off with a general system-prompt that is followed by a specific user-prompt to generate the desired artifact segment. Here it is useful to emphasize the distinction between the "system" prompt and the "user" prompt. The system prompt, shown in Figure 5, provides the model context about the domain (i.e., SE and acquisition of ACAT programs) and the system of interest (i.e., Bulldog, conveyed by a system description and several operational descriptions). Hence the system-prompt's role is to set the context for the LLM prior to any request for a particular SE artifact to be generated.

```
System Prompt: You are a highly skilled Systems Engineer with over 30 years of experience supporting
large-scale Department of Defense acquisitions (including ACAT I, II, and III). In this particular
situation, you are helping write systems engineering (SE) artifacts for the Bull Dog program.

The Bull Dog Unmanned Ground Vehicle (UGV) System will be used in a variety of critical, high-risk
warfighting situations requiring an integrated suite of unmanned air and ground vehicles across
the spectrum of conflict. The Bull Dog UGV System will be integral to the successful
outcome of Reconnaissance, Surveillance, and Target Acquisition (RSTA), mine detection, vehicle
recovery, maneuver, and indirect and direct fire missions, reducing the risk of injury and loss of
life for manned units. Capable of independent operations, the Bull Dog UGV will
carry onboard weapons with the capacity to destroy fixed and mobile targets. The UGV System will
also be capable of line-of-sight (LOS) remote control from the Joint Command and Control System
(JCCS) and non-LOS (NLOS) control via relay using participating UAV assets.

Provided are some operational descriptions:

Interoperability: The Bull Dog must locate and destroy a mobile enemy armored vehicle by using a
laser designator to identify the target so that an AC-130 can conduct the attack. The laser
designator needs to remain on the enemy vehicle for 5-10 seconds to allow communication with the
AC-130 so it can set up the attack.

Mine Detection: The Bull Dog successfully transits an enemy mine field by locating and maneuvering
around the mines. The transit will be unopposed by the enemy.

Seek and Destroy: The Bull Dog must locate and destroy a mobile enemy armored vehicle using only
its onboard weapon systems. The enemy vehicle is hostile and will attack the Bulldog vehicle if it
is detected.

Your task is to support with the generation of systems engineering artifacts. Do not include any
AI-specific language such as "as an AI language model...". Make sure to use DoD-specific language
and terminology to the extent possible. Be verbose in your response in the typical manner of a DoD
systems engineering artifact.

Begin:

Input: {{ Prompt }}
Output:
```

**Fig. 5** The system-prompt used in this study

The system-prompt is then followed by the user-prompt, that is specifically formulated to guide the LLM to generate the expected chunk of the SE artifact. In classical ML terms, this would be the feature engineering requirement required to predict a label (the full text chunk). The user-prompt follows up the system level prompt and is tied to a request for a specific segment of a SE artifact to be generated. Figure 6 presents a user prompt example, generated from the provided capability description document, using a lead question for a particular segment of an SE artifact. The expected output here is a full-text LLM response for each prompt.

```
Input: Based on the information provided, please write me a one paragraph Operational Context
statement for a Capability Description Document. Make sure to mention how it impacts the Joint
Force Commander and what the Bull Dog/UGV System could support in future battlespaces.

Output: The complexity and uncertainty of the current and future Operational Environment (OE)
requires the Joint Force Commander (JFC) to extend the reach of their situational understanding to
continuously adapt to a changing environment across the full spectrum of conflict, while facing
hybrid threats. This extended battlespace awareness requires the collection of information into
the integrated battle command systems network to enable informed decision-making. UGV Systems can
support future forces and expanded battlespace concepts by serving as economy of force assets and
enhancing force protection by providing standoff operational capabilities for many warfighter
functions. UGV Systems must be expeditionary with lethal and non-lethal capabilities that are
versatile, agile, sustainable, survivable, and capable to transition across the spectrum of
operations. UGV Systems must provide persistent surveillance for situational awareness and can
provide force application, including targeting, lethal fires, and non-lethal effects, while
protecting the force through standoff from threat capabilities.
```
**Fig. 6** Example of user-level prompt and the LLM generated output for the given prompt

The final step of our artifact generation approach is the iterative refinement of the user-prompts. We adopt an iterative prompting approach, where we observe the LLM generated response for a given user-level prompt, take it as feedback, and modify our prompt to improve the accuracy of the obtained response. This results in three prompting configurations (for each user-prompt) that increase in terms of their specificity and contextualization. Examples to these prompt configurations are provided in Section 4.1.

Finally, for each SE artifact chunk, we use the system-prompt followed by the 52 user-level prompt pairs (each executed separately) to create responses on the selected LLMs: OpenAI's GPT-4, GPT-3.5 Turbo, and Anthropic Claude[93]. We then proceed into evaluation of these artifacts. Here our goal is not to include all multi-purpose LLMs that are available on the market, but to pick a representative set from the most competent ones that complement our research objective.

Before we proceed to our method for analyzing results, we should remind the reader that this study explores LLM capabilities without the use of explicit fine-tuning or customization. This is a purposeful research choice as there is an ongoing trend in the SE community to use them as such. However, our research design choice also introduces a risk of obtaining over-generalized outputs[94]. To elaborate, given that multi-purpose LLMs are trained using data that is broadly available in the internet; when they are used without fine-tuning, there is a risk of not having enough relevant material in its training corpus, particularly when used for a specialized objective. Hence, its outputs may not be as robust, or well-aligned with the task at hand or user preferences[95,96].

To that end, fine-tuning methods such as full-model tuning[97] and parameter-efficient tuning[98,99] are documented to improve LLM' ability to answer queries regarding domain-specific tasks. However, these methods also have some inherent risks. The former requires extensive storage space, and the latter are influenced by both pretraining and fine-tuning conditions[94]. Additionally, all fine-tuning methods requires sufficient amount of the "right" material to learn from, and could in return, introduce a risk of overfitting[100]. Thus, a practitioner may not be able to spare the resources to garner this increased capability, nor simply be deterred by the associated challenges. Nevertheless, it remains to be seen what is the best approach for generating SE artifacts using LLMs, particularly for open-ended exercises such as problem formulation. This paper takes a step towards that objective, by documenting LLM capabilities for problem formulation tasks without any fine-tuning.

### 3.3. Evaluation of LLM Generated Artifacts
*3.3.1. Quantitative Perspective: Comparison of Text through Natural Language Processing*

The challenge on the evaluation side is the selection of an evaluation framework for comparing a model-generated response to a particular prompt against the human-generated artifact. There are several options at our disposal. BLEU (Bilingual Evaluation Understudy) [101] is typically used for language translation tasks by using n-grams. Recall-Oriented Understudy for Gisting Evaluation (ROUGE) [102] is used for text summarization tasks by comparing the LLM's summary and a reference, human-generated summary. While these evaluation frameworks work well on tasks that have a generally fixed and "correct" output, they are brittle in the context of SE as it is focused on the specific sequences of words that are present, which is a proxy for the actual content of the response. Another alternative is sentence embedding methods such as cosine distances that are commonly used in text analysis particularly in the context of engineering design[103]. These are useful in terms of documenting similarity; however, prioritize frequencies of words and phrases; and could be misleading due to a lack of compositionality in their assessment of a sentence [104].

The third alternative, the MAUVE[49] evaluation aims to "measure the gap between [machine-generated] text and human text" by using Kullback-Leibler (K-L) divergences. A K-L divergence can be thought of calculating the entropy between two probability distributions. In this case, we would produce a distribution of generated text over the "labels" for a given SE artifact (e.g., a natural language conversion of the tabular information), and compare those against the text distribution of what the model generated. Details of the algorithm are described elsewhere[49]; however, in Figure 7 we present a visual representation of how MAUVE proceeds and differentiates Type I and II errors.

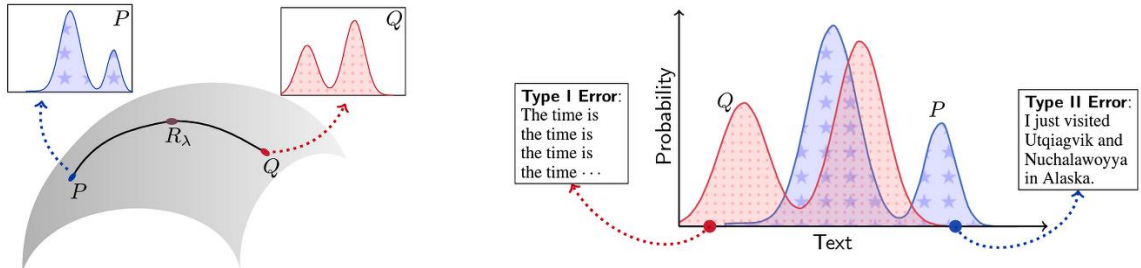

**Fig.7** MAUVE comparison of machine and human text distribution (Image adopted from Pillutla et.al. [49]).

In Figure 7, on the left, MAUVE compares LLM generated distribution *Q* to human generated human text *P*. The right-hand side of Figure 7 visualizes how MAUVE compares *Type I* and *Type II* errors. Here Type I errors refer to a LLM producing degenerative and repetitive text that would be considered unlikely for human generated text. Whereas Type II errors represent expected text segments that cannot be produced by LLM due text due to truncation heuristics[105]. The output of the MAUVE evaluation is a scalar value that ranges between 0 to 1, with 1 meaning that the two distributions are identical (e.g., the LLM was able to output synthetic SE artifacts that directly lined up with what is in Bull Dog). The challenge with the MAUVE is that it requires a significant amount of data, with a minimum of 50 instances and preferably well over 100, but the advantage is it can capture both Type I and Type II errors; and aggregate them for a scalar measure of similarity. Thus, for our evaluation, we will use the MAUVE algorithm to measure the similarities of human vs. AI generated text.

*3.3.2. Qualitative Perspective: Human-Expert Evaluation of AI Generated vs. Human Expert Generated Artifacts*

While quantitative measures provide some insight regarding similarity of two text artifacts, they do not explain *how* these two artifact bodies differ. This is critical for the SE community as artifacts that may appear similar could be missing critical pieces of information or including misleading ones that have to be omitted from further engineering activities. Therefore, we supplemented our investigation with a qualitative analysis of *how* AI generated and human-expert generated SE artifacts differ [106–108]. More specifically, we picked the AI generated SE artifact with the highest MAUVE similarity score, and then contrasted them against the human-expert generated benchmark to understand the nature of deviations. We coded the information based on its informational character and whether it was traceable back to the problem statement conveyed in the prompts.

To ensure consistency of our findings, we followed standard qualitative analysis techniques [109], where we iteratively examined the resulting qualitative codes to extract general features of deviations from the ground truth. More specifically, we had four built-in mechanisms to avoid confirmation bias in our work. First, we decoupled our human-expert benchmark selection from LLM data generation by picking an independently generated SE artifact (Bulldog). This helped to prevent any possible issues regarding what is level of quality would be considered appropriate for the human-expert ground truth. Here, it is useful to note that Bulldog case study was created by a team of SEs, and was controlled by an independent SE team that contracted the associated research project. Therefore, this case selection also allows to control for variability that may originate from an individual human conducting the SE work. Second, we decoupled LLM data-generation process from the qualitative analysis of LLM generated artifacts. The analyzer that created these outputs did not participate in the rest of the data analysis. Third, we used two independent coders for qualitative analysis. Here, while these coders knew they were studying AI generated artifact-prompt pairs, they did not know what other coders were generating. These two coders had diverse educational and professional backgrounds to mitigate any domain specific bias. This information was later synthesized by a third analyzer. Finally, once themes were synthesized, an additional screening was done to seek any counterexamples to these themes. Only the mechanisms that passes this process was labeled and reported in this paper. As discussed further in Section 4.2., this led to identification of mechanisms we label as *failure modes* that characterize how LLM generated artifacts diverge from the human-expert benchmark.

## 4. Findings

We present two main findings through a mixed-methods approach that complements quantitative methods with qualitative ones. First, the ability of multi-purpose LLMs to generate SE artifacts that closely resemble those generated by human-experts. Second, we provide a qualitative explanation of *how* the LLM generated artifacts actually differ from human-expert generated artifacts. In Section 4.1. we present the results of the similarity analysis by using MAUVE, a natural-language processing algorithm. This subsection compares SE artifacts generated by three different LLMs using three prompting configurations that vary in terms of their specificity; against the human-expert generated benchmark. While this analysis enables comparison of large amounts of data, it relies on textual similarities instead of an in-depth quality analysis that necessitates domain-expertise in SE. Therefore, in Section 4.2., we provide the findings from our qualitative analysis, that picks the best LLM generated artifact set, and compares it manually against the human-expert generated benchmark, in search of *how* they differ. This leads to three failure modes through which LLM generated artifacts diverge from human-expert created ones: *premature requirements definition*, *unsubstantiated estimates*, *and propensity to overspecify.*

### 4.1. Quantitative Perspective: Evaluation of Artifact Similarity Using MAUVE

We start our discussion with an evaluation of MAUVE comparisons for LLM generated and human-expert generated artifacts, for all three LLMs included in our study. To recall, MAUVE scores range between 0-1. In this context 1 indicates expert-like and 0 indicates machine generated. Table 1 suggests that results are fairly skewed, and the similarity performance of LLMs to generate expert-like SE artifacts is strongly influenced by prompt engineering. To elaborate, Prompt configurations 1-3 represent user-prompts that range in terms of their specificity; where configuration one is more generic and three requests more specific questions to be answered following certain syntactic requirements. In our case, increasing prompt specificity yielded scores that corresponded to a greater semantic similarity to the expert-generated ground truth. We elaborate these prompt configurations below.

**Table 1** MAUVE Results across 3 LLMs and 3 prompt configurations.

| Model | Prompt Configuration 1 | Prompt Configuration 2 | Prompt Configuration 3 |
|---|---|---|---|
| GPT-4 | 0.0000 | 0.0000 | 0.9137 |
| GPT-3.5 Turbo | 0.0000 | 0.0001 | 0.9749 |
| Claude | 0.0000 | 0.0003 | 0.9932 |

In Prompt Configuration 1, we assigned LLMs a very challenging task of generating a particular SE artifact segment with little direction on the type of content to cover and the suggested length of the response. This led to extremely poor performance – no model was able to generate useful responses and the length of the responses varied wildly. Consider Figure 8 as an example:

```
Input: Based on the information provided, write me the Additional Performance Attribute (APA) for
Rail Transportability including a specific Development Threshold and Development Objective.
Remember you are writing this for a Capability Description Document.
```
**Fig. 8** Example of a User Prompt Configuration 1

This led us to a refinement of our prompting, which provided an indication of improvement in the positive direction, albeit faint. In the Prompt Configuration 2, we edited the user prompts to add more specificity to each instance with regards to the content the LLM would need to generate. Figure 9 presents an example of how the user prompt changed in Configuration 2, for the same SE task. The slight risk with adding specificity is over-generalizing on the provided labels. However, given the low similarity score on the initial run of prompts, this was a risk worth accepting. Nevertheless, we still did not see much change in the MAUVE similarity scores for the LLMs as a result of Prompt Configuration 2, as shown in Table 1.

```
Input: Based on the information provided, write me the Additional Performance Attribute (APA) for
Rail Transportability including a specific Development Threshold and Development Objective. Please
include a very brief rationale on how this requirement benefits the warfighter. Remember you are
writing this for a Capability Description Document.
```
**Fig.9** Example of a User Prompt Configuration 2 for the same SE task in Fig 8.

It was not until the third user prompt configuration that the MAUVE scores significantly escalated. Most surprisingly, the most significant change here was not further refinement of the domain-specific output of each instance, but rather bounding the length of the response. In other words, there was inconsistent language used across the 52 prompt-response instances with regards to the specific length of the expected response. In fact, in one isolated dataset instance, it was observed that the original language asked for a "brief" response, but in absence of more specific criteria (e.g., specific number of sentences or paragraphs), the LLM generated a five-paragraph response. Upon appending each prompt with a specific fixed length based on the associated label/response, the MAUVE scores increased considerably. Figure 10 presents an example of Prompt Configuration 3:

```
Input: Based on the information provided, write me the Additional Performance Attribute (APA) for
Rail Transportability including a specific Development Threshold and Development Objective. Please
include a very brief rationale on how this requirement benefits the warfighter. Remember you are
writing this for a Capability Description Document. Keep the overall response to no more than four
sentences.
```
**Fig. 10** Example of a User Prompt Configuration 3 for the same SE tasks in Figs 8 and 9.

To recap, the introduction of a priori direction in the prompt, which can be proximally treated as having an expert rather than a novice drive the LLM through the generation of SE content, was the key to generating content that most closely resembled the style of the human-expert generated artifacts. What can be inferred here is that LLMs only generate similar answers for those who can evaluate the similarity of the response. A novice SE might ask generic, non-specific questions to an LLM in the course of generating an LLM artifact, which will yield generic, non-specific answers. Only an expert behind the console of an LLM, who has an a priori notion of what the artifact should, and should not, cover in content and in style, would be able to direct the LLM to answer these components. Nevertheless, MAUVE evaluations are only a measure of similarity and do not dictate much on the quality of the content. Thus, we proceed to qualitative evaluation of content.

## 4.2. Qualitative Findings: How LLM Generated Artifacts Differ from the Human-Expert Benchmark

As discussed in Section 4.1, our exercise resulted in nine (3 LLMs and 3 prompt configurations) LLM generated SE artifact sets with 52 user-prompt response pairs in each set. Given the large amount of data, we only picked the LLM generated artifact set with the highest MAUVE score to proceed with the qualitative analysis. In this case it was the Claude generated responses following prompt configuration 3. We present two findings in this subsection. First, we characterized 3 failure modes described below along with some supporting examples. These mechanisms do not always manifest themselves in isolation and are often observed in conjunction with the other failure modes. Nevertheless, we consider it useful to discuss each individually as they may lead to contrasting issues for the rest of the SE activities to follow. Second, we highlight some of the tasks that we observed LLM generated artifacts to closely compare those generated by a human-expert. Although LLMs were not equally consistently good in these tasks, we consider them important to discuss given that they hint at the kind of tasks multi-purpose LLMs might be useful.

*4.2.1. Failure Mode I: Premature Requirement Definition*

This failure mode refers to the inability of multi-purpose LLMs to differentiate needs from requirements, and proceed into a premature and often over-specified definition of requirements, particularly when asked to bound a problem. This failure mode also hints at LLMs struggling to identify the role of the document under consideration in the grand scheme of development activities along with the specific role of the tasks that it is asked to perform; illustrating a lack of SE domain expertise.

For instance, Table 2 shows the results for a prompt-response pair that asks for environmental attributes to be included in a Capability Description Document (CDD). CDDs are created during problem formulation, with the goal to identifying desired system capabilities, and fundamentally includes statements regarding stakeholder needs and objectives[110]. The purpose of a CDD is to serve as an input to definition of system requirements, from which a tradespace analysis and concept exploration will be conducted to identify feasible design solutions. Hence, the role and expectation for CDD is to bound the problem, and not to specify requirements. Our response set included 35 questions similar to this, and we observed in over half of the LLM generated responses, an array of requirement with "shall" statements were returned, without legitimate bounding and support.

Table 2 Example of Failure Mode I: Premature Requirement Definition

| | |
|---|---|
| User Prompt | Based on the information provided, write me a very brief Other System Attribute (OSA) for Environmental. Remember you are writing this for a Capability Description Document. |
| Human-Expert Benchmark | The system must comply with all U.S. environmental regulations and with all international environmental protocols and treaties to which the U.S. is a signatory. |
| LLM Generated Artifact Chunk | The UGV system shall be operable in extreme environmental conditions including temperatures ranging from -20 to +60 degrees Celsius, sand and dust including the ability to withstand sandstorms, as well as conditions of rain, snow, fog and adverse weather. |

Furthermore, we noted three particular issues with this failure mode. First, it usually captured only a subset of relevant factors to consider (in this case temperature, dust etc., among a broader set of relevant environmental factors) without any basis, justification, or traceability. While some of these factors are clearly relevant, it is not an exhaustive list of relevant factors. This is problematic because factor identification at this stage of problem formulation either has to be generally bounded (as it is in the expert benchmark) or has to be exhaustive without any exceptions. Hence, a downstream risk with this failure mode is that if these statements are accepted as-is, they may lead to omission of other environmental attributes that may be relevant. Second, identified variables were stated in an inconsistent level of abstraction. For instance, temperature bounds included precise numerical bounds that were generated without reference to a specific standard, an engineering analysis, or a ConOps. Others factors (e.g., sand, dust, fog) were mentioned without bounds and no information was provided regarding where these bounds could possibly be obtained. Third and finally, some statements included qualitative and vague statements, such as "adverse weather" without specifying what it actually means. In that sense these findings were reminiscent of behavior that would be expected of a novice systems engineer instead of an expert.

On a positive note, we also observed that although the user prompt in Table 2 was fairly vague – as we asked for "Other System Attributes for Environmental", LLM was accurately able to pull information that is in line with the Department of Defense Architectural Framework (DODAF) description that is consistent with the guidance in its system-level prompt. DODAF defines these attributes a category of factors within the operational context that can impact a system's functionality, including geographical location, weather conditions, terrain, electromagnetic interference, and potential environmental threats, all of which need to be considered when designing and deploying a system within a specific environment[111]. Here, although LLM outputs are not identical to the ground truth, it is returning statements that fall into the category of DODAF's definition. We think LLM is actually performing well in terms of accurately sourcing the "right kind" of information; however, it is erroneously executing the prompted task.

In Table 3, we provide another example of premature requirement definition with a different nuance. The LLM is asked to declare a KPP for "Joint Capability", referring to the use case statement in the system prompt regarding collaboration with the A130 system. This is a KPP referring to a conjoined function, which necessitates coordination of interdependent activities, and the ability to relay the information during operations. Hence the expert-benchmark defines two KPPs for this task, one decomposing the target data statement and its bounds in the system-prompt and the second adding an availability KPP referring to collaboration with another supporting system to enable this data transfer. Here, several interesting insights emerge from the LLM response.

Table 3 Premature Requirement Definition: Example of Over-constraining

| | |
|---|---|
| User Prompt | Based on the information provided, write me the KPPs for the Net-Centric Joint Capability, including a specific Development Threshold and Development Objective. Please also provide a very brief rationale. Remember you are writing this for a Capability Description Document. |
| Human-Expert Benchmark | The Key Performance Parameters (KPP) are "dissemination of target data," with a development threshold of <10 seconds and a development objective of <5 seconds; time to establish communications with AC-130 gunship, with a development threshold of <10 seconds and a development objective of <3 seconds, and continuous network connectivity with the Operator Control Unit (OCU), with a development threshold of >0.98 uptime and a development objective of >0.99 uptime. Rationale: Timely and consistent communications and dissemination of target data maximize mission effectiveness and reduce the possibility of fratricide or collateral damage. |
| LLM Generated Artifact Chunk | The KPP for the Net-Centric Joint Capability Area is: The UGV System shall provide a robust net-centric data sharing capability between distributed unmanned platforms to enhance operational battlespace awareness (OBSA) and enable networked force coordination. The Development Threshold is: The UGV System net-centric data sharing capability shall enable two-way exchange of geo-referenced target data, sensor feeds, and situational awareness between a minimum of two unmanned platforms (one ground, one air) operating within line-of-sight (LOS) at a range of 10 km. The Development Objective is: The UGV System net-centric data sharing capability shall enable two-way exchange of geo-referenced target data, sensor feeds, and situational awareness between multiple unmanned platforms (ground and air) operating beyond line-of-sight (BLOS) at a minimum range of 50 km. The rationale is: Net-centric data sharing between distributed unmanned platforms is critical to enhancing OBSA and enabling networked coordinated operations in complex operational environments. |

In Table 3, the first sentence of the response reads like the LLM accurately identifies this task; however, the LLM then proceeds to define a KPP that is reminiscent of a physical boundary, that makes little sense in the scope of the prompt. Although the prompt is asking for a KPP for a "Joint Capability", the response refers to some form of physical distance measured in kilometers that defines an operational boundary. In that aspect, it seems the LLM is *echoing* another segment of the system-prompt, that is not directly related to this specific prompt. Additionally, LLM exhibits an effort to specify "what" this KPP will entail - by using terms such as "geo-referenced target data, sensor feeds" - without articulating "how" this KPP will be measured. Which is a typical premature solution defining behavior. Notice the expert benchmark uses terms purposefully representing inputs and outputs such as "target data"; but does not specify what these interaction terms are going to be composed of. This generic bounding is desirable at the early design phase when a CDD is created, as specifics will be defined at a later stage of development as a function of the solution concept. Finally, the human-expert benchmark includes an explicit consideration of collaborating systems of systems (e.g., A130) and attempts to bound KPPs including this context. This helps establish the boundary of external interfaces and capture directionality of inputs and outputs. Although information about these collaborating systems have been provided to the LLM in the system level prompt (recall Figure 6), there is no information about these operational collaborators in the LLM's response. There were multiple other instances of this failure mode, yet we deem these examples as sufficient and proceed into the next.

*4.2.2. Failure Mode 2: Unsubstantiated Estimates*

The second failure mode is unsubstantiated estimates, where the LLM is asked to provide a bound on some system attributes, and in return provides numerical values that are either completely absurd or possibly conflated with some other characteristic contained in the prompt. Here, similar to the premature requirement definition tendencies, there seems to be a mechanism that seeks information from documents in the training set that is found to be relevant to the topic at hand; and simple values found in these documents are pulled in the response. This is problematic because it completely overhauls the engineering analysis aspect of SE. Thus, the main characteristic of this failure mode is that while LLM does not have the appropriate analytical capabilities at its disposal to be able to generate any of these numerical estimates, it still proceeds to provide some numbers in a palatable form. While we did not expect the LLM to run a lifecycle analysis and calculate what an appropriate unit cost would be; we find it concerning that the LLM readily suggests some numbers instead of arguing it would not be able to execute this task; or at least warn the user about the possible inaccuracies these numbers might exhibit.

Consider the example shown in Table 4, where the LLM is asked to estimate the unit cost for the proposed system of interest, an unmanned ground vehicle. We observe two mechanisms here. First, when the LLM is asked for a unit cost, it returns an exorbitant amount without any support or justification that is more reminiscent of a total system development cost. In fact, the provided unit cost estimate is so significantly off, it dwarfs the most expensive ground vehicles in US Army inventory such as a M1 Abrahams tank (~$15M) or a M104 Paladin ($17.2M)[112]. As such, the provided unit cost estimates even exceed the unit cost of F-35 joint strike fighter[113]. Second, the attribution for rationale is questionable and disregards lifecycle costs on the fleet level[114,115]. Instead of highlighting relevant lifecycle considerations such as total operational costs or replacement costs; and looking for a "right" number of systems to be deployed for desired mission effectiveness, the LLM suggests the proposed unit costs would allow for maximum procurement of the system of interest. Hence although it is not reflected in numbers, there seems to be an underlying cost minimizing argument.

**Table 4** Failure Mode 2: Unsubstantiated Estimates

| User Prompt | Based on the information provided, write me the Key System Attribute (KSA) for System Unit Cost including a specific Development Threshold and Development Objective. Please include a very brief rationale. Remember you are writing this for a Capability Description Document. |
|---|---|
| Human-Expert Benchmark | The UGV System shall have a System Unit Cost of $0.60 M (Threshold)/$0.50 M (Objective). Rationale: System Unit Costs are a key constituent component in reducing TOC and reflect the cost to procure and replace one system. |
| **LLM Generated Artifact Chunk** | The KSA for System Unit Cost for this program is:<br>Development Threshold: $ 383M USD (FY20)<br>Development Objective: $ 351M USD (FY20)<br><br>Rationale: Achieving the Development Objective for System Unit Cost will allow maximum procurement of UGV platforms within budgetary constraints to meet warfighter mission needs. |

Another concerning angle of this failure mode is observed when the LLM is asked a mutually exclusive yet relevant follow-on question and the answers are evaluated in conjunction with each other. In Table 5, the LLM is prompted to provide an estimate on total ownership costs for the system of interest (for the fleet, including acquisition, and operations). In contrast to the unit costs that the LLM provided in Table 4, the LLM significantly underestimates the total ownership cost and even suggests a value that is a fraction of the unit development objective estimate. Numerical estimates in these two examples are not only objectively poor, they also "don't add up" within the given context and are in conflict with one another. Additionally, an aspect of this failure mode we found particularly misleading is the formatting and verbiage in which these numerical estimates are provided. Disregarding the numbers, statements represent the domain jargon and include details that would be expected from an expert, such as specifying fiscal years for time value adjustment or providing threshold and objective values as demanded by the prompt. To summarize, the numerical estimate statements are reasonable in appearance; however, the statements lack the necessary content and coherence.

**Table 5** Unsubstantiated Estimates - Disregard for System Lifecycle

| | |
|---|---|
| **User Prompt** | Based on the information provided, write me the Key System Attribute (KSA) for Total Ownership Cost, including a specific Development Threshold and Development Objective. Please include a very brief rationale. Remember you are writing this for a Capability Description Document. |
| **Human-Expert Benchmark** | The UGV System shall have a Total Ownership Cost (TOC) of $3,500.00 M (Threshold)/ $3,000.00M (Objective).<br><br>Rationale: Total Ownership Cost (TOC) plays a key role in the ability to efficiently field, sustain, and upgrade the portfolio of DoD weapons systems. The requirement reflects the TOC over a 15- year useful life operating in a peacetime scenario. |
| **LLM Generated Artifact Chunk** | The Total Ownership Cost (TOC) KSA for the UGV System shall have a Development Threshold of $25M (Then-Year dollars) and a Development Objective of $15M (Then-Year dollars) in acquisition and sustainment costs from FY2021 to FY2026.<br><br>Minimizing costs through affordable, sustainable, and efficient system design will allow for increased procurement quantity within cost constraints, enabling a broader deployment of this critical warfighting capability. |

*4.2.3. Failure Mode 3: Propensity to Overspecify*

The third failure mode is the propensity to overspecify. While this failure mode appears similar to the first, there are significant differences. Here, the tendency is to introduce additional expectations regarding the system of interest that are often broadly relevant, yet are not needed given the specific context. In other words, the LLM exaggerates the "useful" traits that it thinks the system of interest should exhibit; however, when generating its statements, it loses the anchoring and traceability back to the information provided in the prompts. These overspecifications are not always unreasonable; however, they are rarely correct or necessary, as they are not requested as part of the system or the user prompt. Furthermore, these unjustified statements are often declared with improper bounds that are either unattainable or impossible to verify and validate. Finally, there are issues regarding conciseness, as prompts are often returned as multiple statements packed in one, rather than being decoupled and captured in standalone statements. We illustrate this failure mode with two examples.

In Table 6, the LLM is asked to articulate KPPs for logistics, specifically focusing on deployment, distribution, and maintenance capability, again with threshold and development objectives. Here, we observe that the LLM generated response provides a requirement-like statement that picks an arbitrary "hours of operation without replenishment" metric, for which all logistics characteristics should be built around. The LLM attempts to rationalize the need for this statement, but in doing so, also points out the metrics it should be measured by without establishing a direct link between relevant parameters. It seems the LLM has an understanding of how this system-level metric should be decomposed, yet does not operationalize this general knowledge. This highlights that the LLM does not necessarily "know" the subject when information is linked and presented to the requester; rather, the LLM simply presents knowledge that the LLM deems relevant in a conversational way.

**Table 6** *Failure Mode 3: Propensity to Overspecify – Introduction of Unjustified Variables*

| | |
|---|---|
| **User Prompt** | Based on the information provided, write me the KPP for the Logistics (Deployment and Distribution, and Maintain) Joint Capability Area, including a specific Development Threshold and Development Objective. Please also provide a one paragraph rationale. Remember you are writing this for a Capability Description Document. |
| **Human-Expert Benchmark** | The Key Performance Parameter (KPP) is the operational availability, with a development threshold of >0.80 uptime and a development objective of >0.95 uptime. <br><br> Rationale: The UGV shall be employed in various terrains and environments. The UGV will be supported over its life cycle by replacing individual parts as they wear out. The system is expected to be in service for a period of 15 years and shall be designed for a minimum service life of 15 years. It can be calculated using logistics parameters that consider the effect of reliability, maintainability, and availability. <br><br> The attributes to calculate are the Mean Time Between Failure (MTBF), Mean Time To Repair (MTTR) and the Mean Logistics Delay Time (MLDT). Operational Availability = MTBF/(MTBF+ MTTR + MLDT). It is the quantitative link between readiness objectives and supportability. |
| **LLM Generated Artifact Chunk** | The UGV system shall be capable of deploying, distributing itself for operation, and sustaining operations at a threshold minimum of 96 hours without replenishment and an objective minimum of 336 hours without replenishment. <br><br> Rationale: To meet operational requirements across a dispersed operational environment, the UGV system must have a high mean time between failure (MTBF), low mean time to repair (MTTR), and low mean logistical downtime (MLDT). A 336-hour sustainment capability without replenishment at the Development Objective will ensure that the system can operate for extended periods across a range of operational tempos in austere and contested environments |

| | without reliance on vulnerable logistical supply lines. The 96-hour threshold will provide for a minimum self-sustaining capability to meet critical missions while balancing cost and technical risk at incrementally increasing capability over time. |
|---|---|

Another example is provided in Table 7, where the LLM is requested to generate a KPP for force protection, which in this context corresponds to detection of landmines, the primary functionality of the system of interest. Here, the LLM not only provides somewhat of an irrelevant statement, but has also sets forth a KPP that is impossible to verify or attain. In this particular case, this is reflected in two aspects. First, inclusion of an array of stakeholders, some of which are unrelated to the information passed to the LLM in prompts. Second, expression of an unattainable development objective as a KPP, that is captured with the statement "shall mitigate loss of **all…**"; another overly ambitious and un-verifiable objective.

**Table 7** *Propensity to Overspecify – Introduction of Unattainable or Verifiable Bounds*

| User Prompt | Based on the information provided, write me the KPP for the Force Protection (Prevent and Mitigate) Joint Capability Area, including a specific Development Threshold and Development Objective. Please also provide a very brief rationale. Remember you are writing this for a Capability Description Document. |
|---|---|
| Human-Expert Benchmark | The Key Performance Parameter (KPP) is the probability of detection for mines, with a development threshold of >0.95 likelihood and a development objective of >0.99 likelihood.<br><br>Rationale: The capability for the UGV System to detect mines and disseminate mine location data enhances mission effectiveness and reduces injury/loss of life and damage to equipment. |
| LLM Generated Artifact Chunk | KPP for Force Protection (Prevent and Mitigate): The UGV System shall provide scalable levels of force protection for personnel, systems, infrastructure, and critical assets to prevent and mitigate loss of Soldiers, civilians, infrastructure, materiel, and critical information through enemy attack, terrorism, accidents, natural disasters, disease, and exposure.<br><br>Development Threshold (Min): The UGV System shall mitigate loss of all protected items to a level commensurate with CONPLAN standards.<br><br>Development Objective (Optimal): The UGV System shall prevent and mitigate loss of all protected items to a level that enables continuance of operations and continuity of government and services.<br><br>Rationale: This KPP establishes the force protection, prevent and mitigate capability UGV System performance requirements to ensure protected items are suitably secure from loss, with the aim of achieving successful continuance of operations and continuity of government. |

A plausible explanation to why this failure mode might be occurring could be based on the mechanism the LLM gathers and generates this information. To elaborate, recall that the LLM is provided with a system level-prompt that provides the general context and a user-level prompt that asks for a specific task. It seems the LLM processes the user-prompt and then seeks some anchoring segments of text in the system prompt. These anchors might be used to search within its pre-training corpus to find similar artifacts that contain similar statements, from which plausibly relevant terms are scraped, converted, and provided as a response. These often include arguments about the given prompt, yet are irrelevant in terms of the context and mission of the system of interest as summarized in the prompts. We more frequently observed this failure mode for performance related questions; however, it was manifested in other user-prompts as well.

*4.2.4. Satisfactory Traits*

We discovered some positive trending traits of the LLM that we highlight in this section. Nevertheless, these were not as consistently revealed in our exploratory analysis. Thus, we urge the readers to consider these as sporadic indicators of some cases in which multi-purpose LLMs could prove useful for generating SE artifacts, and not in terms of generalizable patterns. These could be summarized with: (i) the ability to follow guidance, particularly in terms of formatting, such as in providing development thresholds and objectives; (ii) adjustment of information as it would be expected from expert, such as in the case of Tables 4 and 5 with the adjustment for fiscal year dollars; and (iii) successfully summarizing context related statements.

The responses do not seem to be stemming from a common aspect in the prompted text from either the style of prompt or specific content that is attempting to be ascertained. Consider Table 8, where the LLM provides a potentially viable answer for a context description task that only slightly differs from the human-expert benchmark in terms of the focus of the response. Here, the human-expert benchmark focuses on general purpose of the system of interest while the LLM focuses on its autonomous nature, and implicitly, how that relates to the purpose. Nevertheless, they are not identical or arguably on the same level of quality. For instance, the LLM response focuses on emphasizing "how" the system of interest will be providing some of these mission level capabilities (e.g., "with onboard sensors and weapons…"). Beyond this difference of perspective both SE artifacts in Table 8 arguably provide similar information and could be considered as passable. Variable content tends towards the LLM response reading like a human-expert writer with different preferences.

**Table 8** Example of a similar LLM generated response in comparison to the Human-Expert Benchmark: Perspective Shifts

| User Prompt | Based on the information provided, please write me a one paragraph Operational Context statement for a Capability Description Document. Make sure to mention how it impacts the Joint Force Commander and what the Bull Dog/UGV System could support in future battlespaces. |
|---|---|
| Human-Expert Benchmark | The complexity and uncertainty of the current and future Operational Environment (OE) requires the Joint Force Commander (JFC) to extend the reach of their situational understanding to continuously adapt to a changing environment across the full spectrum of conflict, while facing hybrid threats. This extended battlespace awareness requires the collection of information into the integrated battle command systems network to enable informed decision-making. UGV Systems can support future forces and expanded battlespace concepts by serving as economy of force assets and enhancing force protection by providing standoff operational capabilities for many warfighter functions. UGV Systems must be expeditionary with lethal and non-lethal capabilities that are versatile, agile, sustainable, survivable, and capable to transition across the spectrum of operations. UGV Systems must provide persistent surveillance for situational awareness and can provide force application, including targeting, lethal fires, and non-lethal effects, while protecting the force through standoff from threat capabilities. |

| LLM Generated Artifact Chunk | The UGV system will provide capability to the Joint Force Commander enabling the extended battlespace awareness and freedom of maneuver necessary to decisively defeat adversaries across the spectrum of conflict. By reducing force structure requirements for critical yet high-risk warfighting functions, the UGV system allows the Joint Force Commander to employ a smaller force footprint while retaining operational dominance. Specifically, the UGV system can support economy of force RSTA, maneuver, fire support, and force protection missions, allowing the Joint Force Commander to allocate a higher percentage of manned forces to decisive operations. With onboard sensors and weapons, the UGV system provides 24/7 persistent stare and precision engagement to detect, identify, and defeat threats within the extended battlespace. The UGV system's ability to conduct autonomous lethal and non-lethal effects and seamlessly interface with manned and unmanned systems empower the Joint Force Commander with flexible, scalable, and layered options for shaping the extended battlespace. |
|---|---|

There were other examples where we found the content of LLM generated artifacts to be acceptable; however, these were often accompanied by an unnecessary level of detail that may or may not be essential given the level of analysis. In Table 9, the task is to define attributes for transportation to the theater. Here the human-expert benchmark states the external boundary and the transportation mode, by declaring the system should be packed inside a standard shipping container. Instead, the LLM response focuses on modes of transportation (e.g., air, rail) and sets to articulate what kinds of supporting systems may be conducting these tasks. While, inclusion of these external systems may be considered undesirable and over constraining for a CDD; we found the manner in which LLM provided this information useful, which reads more as a suggestion and not as a premature requirement. Identification of these external collaborating systems may potentially allow the designers to more explicitly consider relevant performance and functional characteristics that originate from these choices, such as vibration profiles and shocks[116]. Upfront consideration of boundary conditions may enable more proactive decision-making in earlier stages of design.

Table 9 Example of a similar LLM generated response in comparison to the Human-Expert Benchmark

| User Prompt | Based on the information provided, write me a very brief Other System Attribute (OSA) for Transportation to the theater. Remember you are writing this for a Capability Description Document. |
|---|---|
| Human-Expert Benchmark | The UGV System shall be capable of transportation to the theater of operation in standard shipping containers. These containers shall be selected from the existing government inventory. Once in theater, the UGV System shall be capable of rapid deployment to forward operating areas. |
| LLM Generated Artifact Chunk | The UGV System must be capable of being transported by sea (via high-speed sealift), air (C-17 Globemaster III or larger aircraft), road (via heavy equipment transporter) or rail to a forward operating base, then transported via UGV transporter or sling loaded under a heavy lift helicopter (CH-53K or larger) over extended ranges/distances to reach its operational area. |

## 5. Discussion

Advancing capabilities in generative AI, particularly evidence from multi-purpose LLMs passing domain-expertise exams in the fields of law[117] and medicine[118,119]; is encouraging the SE community. However, SE problems are interdisciplinary, complex, and highly context dependent. Translating ill-defined SE problems to well-defined ones that are tractable for solving[13,19,20] requires knowledge from multiple complementary domains to be synthesized within the specific context of the given system of interest. To that end, this study adopted a mixed-methods approach to investigate the extent to which multi-purpose LLMs can generate expert-like SE artifacts; and documented differences and similarities. As illustrated in this study, while multi-purpose LLMs can generate segments of SE artifacts that closely resemble those generated by human experts an in-depth evaluation of these artifacts reveals concerning trends regarding content. As a preface to this discussion, we would like to remind the reader that between the time this study was conducted and published, LLMs investigated here have already been replaced by more capable ones in terms of general performance. Nevertheless, these findings document the state of the art at the time for a SE problem formulation task, and this paper could serve as a benchmark for future work. This study makes three important contributions.

First, although exploratory in its nature given that we only worked with a very limited set of benchmark SE artifacts and prompting approaches, findings reported in this study provide the existence proof for the ability of multi-purpose LLMs to assist with SE tasks, at least for a series of problem formulation use cases. While we documented that LLMs can generate expert-like SE artifacts without any explicit fine-tuning, this closeness refers to semantic similarity rather than quality. We found no significant difference between different LLMs explored in this study regarding this aspect. However, Claude performed marginally better in terms of similarity, possibly due to its relatively high input token limit, followed by GPT-3.5 Turbo and GPT-4. Here, we should remind the audience that this relative ranking pertains to semantic similarity for this particular case study, and does not suggest a general ranking of quality. Nevertheless, prompt engineering plays an influential role in the closeness of artifacts, at least when using the MAUVE framework for measurement. The low similarity scores observed in earlier prompt configurations were due to the MAUVE evaluation framework was detecting large Type II discrepancies between human-generated and LLM generated responses due to high variability. As soon as that was controlled for by refinement of the prompts, LLMs fared far more capable. To that end, while prompt engineering may not be the future for how generative AI could be best utilized for SE, it could help save the day[67]. Related to this finding, we anticipate that newer, more capable LLMs will perform significantly better.

Second, through an in-depth qualitative inductive approach, we documented three failure modes that explain how AI generated SE artifacts may differ from human-expert generated ones. We labeled these failure modes as premature requirements definition, unsubstantiated estimates, and propensity to overspecify. Each of these bear the risk of introducing significantly misleading information into the design process or omitting essential ones. If taken for granted and not examined rigorously, these failure modes could introduce cascading errors into the design process that may lead to over-design, omission of relevant information, and ultimately improper bounding of design problems, particularly for the problem formulation use-case we explored. What is more concerning is that these statements are often posed in a semantic format that "looks good", resembling information that would be expected from an expert.

What can be said about the shared patterns between these failure modes? In terms of generalizability, although specifics of these failure modes originate within the context of the system of interest, resulting insights are generic and likely transferable to similar LLM applications regarding SE and other similar disciplines where domain-relevant knowledge may not be available in the training corpus. Related to this point, identified failure modes are reminiscent of those that would be expected from a novice systems engineer or a disciplinary engineer trying to serve as a SE. One could argue conflating user needs and system requirements, over-constraining a design space, or providing poor numerical estimates, are indeed not unique to LLMs and are frequently committed errors by humans. Nevertheless, these are novice mistakes that are not expected of experts – as accurately reflected in the benchmark used in this study. Regardless, nature of the tasks explored in this study exploited the lack of engineering analysis capability of LLMs, along with the lack of deep contextual knowledge, suggesting that multi-purpose LLMs could continue to struggle with SE relevant tasks.

In terms of style, overall, LLMs provided more verbose responses than the benchmark, where the line between useful and useless arguments are hard to navigate. The lengthy response can likely be attributed to the LLM presenting broad information of similar sources for the narrow scope of the specific prompt. In some cases, this leads to irrelevant responses for the system of interest and in others sweeping sets of arguments that attempt to cover irrelevant characteristics. Thus, these observations support our claims of an inability to infuse expert judgement, and "echoing back themes" that LLM finds plausible. Similarly, we found that vague prompts or prompts that might require deep domain knowledge, were challenging LLMs given their lack of intuition for the knowledge. A potential remedy to this problem could be leveraging rule and knowledge based approaches[39,40,83,50], or LLMs that are fine-tuned on proprietary datasets that somewhat capture desired quality threshold[44,50]. There are an array fine tuning methods discussed in Section 3.2 could be leveraged for this purpose in future work. Ultimately these responses suggested a continued need for the guidance of a subject matter expert when working with AI[88,86].

Third, we documented certain tasks for which LLMs fared relatively well. These can be summarized with the ability to follow guidance on formatting, presentation of information as it would be expected from an expert, and efficient summarization of broadly relevant concepts regarding the context. Neither of these tasks require SE expertise and at best should be considered as an assistant reminding useful information. This is illustrated not only in the overlap between the responses but for the model's ability to disambiguate acronyms when placed in a specific domain or context. Nevertheless, our findings should only be treated as a very conservative demonstration of what multi-purpose LLMs can achieve for SE; and a potential path forward should arguably make use of both more relevant training material and expert knowledge. Given that many SE organizations have access to enormous domain-specific datasets, generating and utilizing similar generative models in their own repositories may prove more effective. There are limited examples to this[50], but they should be considered as trailblazers of a new era.

An important aspect to emphasize before we conclude this paper is that the task of problem formulation may not be a very good application of LLMs, particularly without fine-tuning for a very specialized domain like SE for defense systems and perhaps for specialized product such as a CDD. Here, we do not intend to claim that many SE organizations are expecting to use LLMs for this kind of open-ended exercises. More suitable applications, particularly without the availability of a large fine-tuning corpus of relevant domain-specific data may include text summarization, text evaluation against specific quality or content criteria, reframing of requirements into structured requirement patterns, editorial cleanup, consistency checking, etc. We discuss some of recent research in Section 2.2.; however, this is not an exhaustive list and there is plenty of room for future research. Additionally, if a specific sector (e.g., US Department of Defense, civil aviation, space agencies) aims to leverage these technologies more broadly, it would be worthwhile to invest in specializing the LLM for the specific sector through methods such as fine-tuning based on creation of a large corpus of training material. This could propel a new era of LLMs; however, it remains to be seen how such data could be shared without compromising any sensitive information. Below, we delve deeper into how far we expect these findings to hold, limitations of this work, and some suggestions for future research.

### 5.1. Expectations Regarding Future Relevance of Findings

LLMs are evolving at an astonishing pace. As such, between the time this study was conducted and published, the LLMs examined in this study have been already replaced by newer versions with considerable general performance improvements. While this paper does not intend to provide an evaluation of the limits of LLM technology in general, we contend that it is useful to discuss how far we anticipate our findings, particularly those that refer to the ability of future LLMs to address the identified *failure modes*, will hold.

We expect that *unsubstantiated estimates* and *propensity to overspecify* might be difficult to counter, at least in terms of providing high quality answers. This is because problem formulation is a challenging SE activity that requires substantial domain-expertise given the open-ended nature of translating an ill-defined problem[120,19,20], often-unique contextual environment, and the need to cater to the counterbalancing objectives of the system of interest. Thus, unless there is a fundamental change in how LLMs operate, looking across other document bodies to identify relevant information may not be a viable approach for SE problems, at least for systems with sufficient complexity or uniqueness. This may be less true for more standardized and relatively incremental products such as basic consumer products and automobiles, compared to one-of a kind, cutting-edge systems such as the James Webb Telescope[5]. Alternatively, expanding LLMs with dedicated engineering knowledge-based expert models, such as expert systems or its hybrids, could be help navigate these failure modes.

Finally, while we have reasonable evidence to suspect *premature requirements definition* will persist for a foreseeable future similar to the previous two failure modes, it may be marginally easier to counter by future LLMs; if they could correctly recognize that the query is not to define a requirement but is to bound the problem. This is often an elusive challenge for novice SEs and that may also be the case for LLMs. Here a viable improvement could be guiding the user towards relative mission and context information, or synthesizing some of this information to inform their work. In other words, even if future LLMs could be trained to declare "what they don't know" it would be a significant improvement.

### 5.2. Limitations

We should also highlight some of the limitations of this study. An important research design decision we made was to structure every prompt as a silo supported by a system-prompt. Thus, generated chunks of SE artifacts did not refer to itself. This was a purposeful decision to balance the token limits of each LLM and to avoid "cascading hallucinations" that may originate from the human-in-the-loop attempting to push the LLM to explain and link things back to its previous responses that may or may not be directly relevant. Possibly as a consequence of this, identified failure modes didn't necessarily reflect much of hallucinating[121,122]. This could also be partially a result of basing our qualitative analysis on Claude, which are documented to be relatively better in terms of lack of hallucination, compared to other LLMs.

Another research choice that is important to note was that our experimental procedure was not blinded, meaning that while two independent coders were employed to execute the qualitative coding and they knew they were analyzing LLM outputs. Here, the possibility of introducing confirmation bias in the findings was countered by relying on a third coder to aggregate the qualitative coding into the failure modes. Thus, while we believe it is not a threat to validity, it is useful to note for replicability of results.

Additionally, we should note that LLMs are not deterministic tools, and every instance of LLM usage with identical inputs may lead to different results. This study did not aim to measure the stability or variance of LLM generated responses, and rather assumed LLMs would behave in a rather deterministic manner. This is a strong assumption that may be explored later, perhaps in a simulated experiment[123]. Related to these challenges, future research could explore how the multi-purpose LLMs fare for various SE tasks, how much variability is contained within the AI generated response, and may test the sensitivity of documented failure modes following different prompting approaches.

### 5.3. Some Future Research Opportunities

Our study also illuminates some research opportunities for improving the effectiveness of LLMs in the context of SE. The most pressing are:
- **Limits of available training data:** while some SE relics are available in the public pre-training corpus, most Defense-specific acquisition information is not publicly released and is stored in either classified or proprietary networks. This means that multi-purpose pre-trained LLMs such as the ones we investigated in this study are fundamentally handicapped to producing content that resembles what is commercially available. Prompt engineering can make up some of the gap but is not a replacement for an enterprise-grade fine-tuning strategy. Nevertheless, this does not prevent organizations from developing their own tools, that can utilize both publicly available information and their own organizational knowledge bases, which could help alleviate this limitation.

- **Risk of over-engineering model outputs:** we saw that the model outputs are significantly more valuable when the prompt is engineered towards specific responses. However, this concurrently introduces a risk that model responses will produce content exactly like previous acquisitions were designed, which may be undesirable compared to a situation where the model is producing useful and novel ideas on system design. As an extreme example, it is not desirable for SE artifacts to be produced on the basis of the original B-52 aircraft when the new system capability is expected to resemble that of the F-35 aircraft and, vice versa, it is not desirable for SE artifacts to be produced on the basis of an F-35 when the need can be met with the simplicity of the original B-52.
- **Trust and reliability:** LLMs are known to be susceptible to vulnerabilities either due to the nature of their training (e.g., as discussed in [90]) or with biases inadvertently introduced during the SFT or RL phases. Thus, at least for a foreseeable future, there should be a human expert-in-the-loop to audit model responses to ensure quality. This leads to the corollary that there is significant room for research on human-AI collaboration on both development and verification of trustworthy LLMs.
- **Performance gaps between open- and closed-source models and the inherent security risks associated with using them:** it is observed that closed-source models, generally created by larger, well-funded corporations who can pay for larger datasets and very expensive SFT datasets, generally perform larger than open-source models. For most enterprise applications, closed-source models are infeasible as sensitive data would have to hit commercial servers to allow for model inference. While the results presented in this article suggest value in closed-source models for generation of SE artifacts, a more comprehensive evaluation should be completed across the full ecosystem of LLMs. Nevertheless, to the best of our knowledge, research on the possible scale, and implications of, associated security risks is nascent.
- **Consistency and Stochastic Nature of LLMs:** Although we ignored this fact in this study, LLMs are not deterministic tools. Meaning that for open-ended exercises, such as SE problem formulation tasks, they may provide different answers in each run. Research is needed to test and monitor consistency of these tools, with an explicit understanding of the quality of their outputs. Natural language processing techniques could be leveraged to analyze these results in batch; however, it is still a significant research challenge. We contend that applicability of LLMs for SE purposes will hinge on their ability to navigate this challenge.

## 6. Conclusions

To conclude, LLMs are remarkable tools, having become adopted across government and commercial sectors in a matter of months and become global household names. It is highly likely that in a matter of years, they will be adopted as a co-pilot tool to either generate artifacts, provide some level of critiquing on existing information, support information retrieval from MBSE tooling, apply new technical frameworks to existing or on-going program acquisition, etc. Nevertheless, this paper illustrates that "commercial-off-the-shelf" value of LLMs for SE tasks is rather limited at this stage, with engrained failure modes that may lead to poor SE decisions. There is significant room for research on methods and techniques to improve both infusion of expert knowledge and utilization of multi-modal design information to make it a true enterprise-grade solution.

**ACKNOWLEDGMENT**

An earlier version of this research was first discussed at Conference on Systems Engineering Research (CSER) 2024 in Tucson, AZ and was the recipient of the Best Paper Award[48]. This article significantly differs from the conference version with the addition of a more elaborate literature review, the qualitative analysis, and the categorization of LLM failure modes.

This material is based upon work supported, in part, by the U.S. Department of Navy, Navy Engineering Education Consortium (NEEC), under Contract [N00174-22-1-0021]. Any views, opinions, findings and conclusions or recommendations expressed in this material are those of the author(s) and do not necessarily reflect the views of the United States Department of Navy.

**CONFLICT OF INTEREST STATEMENT**

The authors have no conflicts of interest to declare.